\documentclass[utf8]{FrontiersinVancouver} 
\usepackage{url,hyperref,lineno,microtype,subcaption}
\usepackage[onehalfspacing]{setspace}

\def\keyFont{\fontsize{8}{11}\helveticabold }
\def\firstAuthorLast{Bigolin Lanfredi} 
\def\Authors{Ricardo Bigolin Lanfredi\,$^{1,*}$, Joyce D. Schroeder\,$^{2}$ and Tolga Tasdizen\,$^{1}$}
\def\Address{$^{1}$Scientific Computing and Imaging Institute, University of Utah, Salt Lake City, UT, USA\\
$^{2}$Department of Radiology and Imaging Sciences, University of Utah, Salt Lake City, UT, USA}
\def\corrAuthor{Ricardo Bigolin Lanfredi}

\def\corrEmail{ricbl@sci.utah.edu}

\usepackage{amsfonts}
\usepackage{dsfont}
\usepackage[export]{adjustbox}
\usepackage[acronyms]{glossaries}
\usepackage{float}

\newacronym{CNN}{CNN}{convolutional neural network}
\newacronym{CXR}{CXR}{chest x-ray}
\newacronym{ET}{ET}{eye-tracking}	
\newacronym{AUC}{AUC}{area under the receiver operating characteristic curve}	\newacronym{CAM}{CAM}{class activation map}
\newacronym{IoU}{IoU}{intersection over union}
\newacronym{GT}{GT}{ground truth}

\begin{document}
\onecolumn
\firstpage{1}

\title[Eye-tracking annotation for label-specific localization]{Localization supervision of chest x-ray classifiers using label-specific eye-tracking annotation} 

\author[\firstAuthorLast ]{\Authors} 
\address{} 
\correspondance{} 

\extraAuth{}

\maketitle

\begin{abstract}

\section{}
\Glspl*{CNN} have been successfully applied to \gls*{CXR} images. Moreover, annotated bounding boxes have been shown to improve the interpretability of a \gls*{CNN} in terms of localizing abnormalities. However, only a few relatively small \gls*{CXR} datasets containing bounding boxes are available, and collecting them is very costly. Opportunely, \gls*{ET} data can be collected in a non-intrusive way during the clinical workflow of a radiologist. We use \gls*{ET} data recorded from radiologists while dictating \gls*{CXR} reports to train \glspl*{CNN}. We extract snippets from the \gls*{ET} data by associating them with the dictation of keywords and use them to supervise the localization of specific abnormalities. We show that this method improves a model's interpretability without impacting its image-level classification. 

\tiny
 \keyFont{ \section{Keywords:} eye tracking, chest x-ray, interpretability, annotation, localization, gaze} 
\end{abstract}
\glsresetall
\section{Introduction}

Along with the success of deep learning methods in medical image analysis, interpretability methods have been used to validate that models are working as expected~\citep{chexnet}. The interpretability of deep learning models has also been listed as a critical research priority for artificial intelligence in medical imaging~\citep{roadmap}. The employment of bounding box annotations during training has been shown to improve a model's ability to highlight abnormalities and, consequently, their interpretability~\citep{cvpr2018}. However, bounding boxes for medical images are costly to acquire since they require expert annotation, whereas image-level labels can readily be extracted from radiology reports. This fact is exemplified by the relatively small size of bounding box datasets for \glspl*{CXR}~\citep{vindrcxr,chestxray8} when compared to the size of \gls*{CXR} datasets with image-level labels. \Gls*{ET} data, on the other hand, contain implicit information about the location of labels, and its collection can potentially be scaled up if the acquisition of gaze from radiologists is implemented in clinical practice.

\Glspl*{CXR} are the most common medical imaging exam in the United States~\citep{mostcommon}. This type of imaging has also had much attention from deep learning practitioners, with successful technical results~\citep{chexnet,radreview}. Despite their universality, reading a \gls*{CXR} is considered one of the hardest interpretations performed by radiologists~\citep{hard}, with high inter-rater variability in reported abnormalities~\citep{lowrater1,lowrater2}. Moreover, abnormalities in \glspl*{CXR} can appear in a vast diversity of locations, including the lungs, mediastinum, pleural space, vessels, airways, and ribs~\citep{padchest}. They can also be described as hundreds of different findings~\citep{padchest}. The diverse aspect of the report has been simplified for use in deep learning applications, where, in several cases, a simplified subset of the most common labels has been automatically extracted from reports for use in a multi-label formulation~\citep{chexpert,chexnet,mimiccxr}. Since radiologists pay close attention to several areas when dictating a \gls*{CXR} report, scanning almost the whole image for signs of several abnormalities, the \gls*{ET} data accumulated during the full report dictation might highlight several areas with no evidence of abnormality. Therefore, the use of the temporal aspect of the report, by processing the \gls*{ET} data with the dictation-transcription timestamps, is expected to achieve a more precise localization for specific abnormalities.

We propose to use \Gls*{ET} data with timestamped dictations of radiology reports to identify when the presence of specific abnormalities was dictated and extract the associated gaze locations. The extracted information can be used as label-specific annotation for supervising models to highlight abnormalities spatially. The localization supervision is performed using a combination of a multiple instance learning loss over the last spatial layer of a \gls*{CNN}~\citep{cvpr2018} and a multi-task learning loss~\citep{ibm}, adding an output representing label-specific \gls*{ET} maps. To complement the annotations of an \gls*{ET} dataset, we employ a large dataset of \glspl*{CXR} with image-level labels in a weak supervision formulation. We evaluate the classification performance and the ability to localize abnormalities of a model trained with data annotated by \gls*{ET}. This model is compared against baselines using no annotated data and hand-annotated data. We show that using \gls*{ET} data during training improves the localization performance of generated interpretable heatmaps without compromising \gls*{AUC} classification scores and that this type of data can potentially replace hand-labeling. In addition, to the best of our knowledge, this study offers the first estimation of how the value of \gls*{ET} data compares to the value of hand-annotated localization data when a very large dataset with image-levels labels is available for weak supervision.

\section{Materials and methods}

\subsection{Extracting localization information from eye-tracking data}
\label{sec:extraction}

We designed a pipeline to extract disease locations from \gls*{ET} data. This pipeline has two main parts: extracting label mentions in reports and generating an \gls*{ET} heatmap for a given detected label. A representation of the pipeline is shown in Figure~\ref{fig:method}. The pipeline requires the \gls*{ET} dataset to contain timestamps, transcriptions of report dictations, and fixations, i.e., locations in the image where radiologists stabilized their gaze for some time.

To extract labels from reports, we adopted a modified version of the CheXpert labeler~\citep{chexpert}, which uses a set of hand-crafted rules to detect label mentions and negation structures. 

From observing the gaze of radiologists on a few examples, we noticed patterns that seemed to be in common for all radiologists:
\begin{itemize}
\item for the first moments after being shown a \gls*{CXR}, radiologists looked all over the image without dictating anything;
\item when dictating, radiologists usually looked at regions corresponding to the content of the current sentence or the following sentence (when near the end of the dictation of the current sentence).
\end{itemize}
From these two observations, we decided to generate \gls*{ET} heatmaps for detected labels from the fixations of the sentences where the label was mentioned, the previous sentence, and the pause between sentences. We accumulated fixations within a limit of 1.5 seconds previous to the start of the mentioning sentence and up to the last mention in the mentioning sentence. An illustration of the method for choosing which fixations were included in the heatmaps is given in Figure~\ref{fig:15smethod}. An example with a case from the \gls*{ET} data we used are given in Figure~\ref{fig:method}a and~\ref{fig:method}b. Heatmaps were generated by placing Gaussians over each fixation location with a standard deviation of one degree of visual angle, following Le Meur et al.~\citep{onedegree}. Fixations had the amplitude of their Gaussians weighted by their duration. The heatmap for each detected mention of a label was normalized to have a maximum value of 1. Multiple mentions of a label for the same \gls*{CXR} were aggregated with a maximum function.

\subsection{Multiple instance learning}
\label{sec:lietal}
We used the multiple instance learning loss term from Li et al.~\citep{cvpr2018} to train an encoder with the extracted \gls*{ET} heatmaps. The encoder $E$, as represented in Figure~\ref{fig:method}c, was built to output a grid of cells, where each cell represented a multi-label classifier for label presence in the homologous region of the image. The image-level prediction $C_k(x)$ for label $k$ was formulated as

\begin{equation}
\label{eq:output}
C_k(x) = 1 - \prod_{j\in \Gamma_x}{1 - \sigma\left(\gamma_{jkx}\right)},
\end{equation}
where $\gamma_{jkx}$ is the logit output for class $k$ and grid cell $j$ for image $x$, $\Gamma_x$ is the set of all grid cells for image $x$, and $\sigma(\cdot)$ is the sigmoid function. Equation~\ref{eq:output} is a soft version of the Boolean \textit{OR} function, assigning a positive image-level label when at least one of the grid cells was found to contain that class. 

During training, the loss function depended on the presence of a localization annotation. For images annotated with localization (A), grid cells were trained to match a resized version of the annotation, as shown in Figure~\ref{fig:method}c. We used the loss
\begin{equation}
\label{eq:lossannotated}
L_{kA}(x)= -\text{log}\left(\prod_{j\in B_{kx}}{\sigma\left(\gamma_{jkx}\right)} \prod_{j\in \Gamma_x - B_{kx}}{1 - \sigma\left(\gamma_{jkx}\right)}\right),
\end{equation}
where $B_{kx}$ is the set of grid cells labeled as containing evidence of disease $k$ for image $x$ and $L_{kA}(x)$ is the output of the loss function for annotated images of class $k$.

For images that did not contain localization annotations (U), the loss depended on the image-level label. For positive images, at least one grid cell should be positive. We used the loss
\begin{equation}
\label{eq:lossuplus}
L_{kU^+}(x) = -\text{log}\left(C_k(x)\right),
\end{equation}
where $L_{kU^+}(x)$ is the output of the loss function for unannotated images labeled as positive for class $k$. For negative images, all grid cells should be negative. We used
\begin{equation}
\label{eq:lossuminus}
L_{kU^-}(x) = -\text{log}\left(\prod_{j\in \Gamma_x}{1-\sigma\left(\gamma_{jkx}\right)}\right)
\end{equation}
where $L_{kU^-}(x)$ is the output of the loss function for unannotated images labeled as negative for class $k$. The multiple instance learning loss term $L_{I}(x)$ was then formulated as 
\begin{equation}
\label{eq:lossmil}
L_{I}(x) = \mathbb{E}_{x \in X,\,k \in K}\left[{\lambda_{A}\mathds{1}_{kA}(x) L_{kA}(x) + \mathds{1}_{kU^{+}}(x) L_{kU^{+}}(x) + \mathds{1}_{kU^{-}}(x) L_{kU^{-}}(x)}\right],
\end{equation}
where $\lambda_{A}$ is a hyperparameter controlling the relative importance of annotated images during training, $X$ is the set of all images, annotated and unannotated, $K$ is the set of all classes, and $\mathds{1}_{kA}(x)$, $\mathds{1}_{kU^+}(x)$, and $\mathds{1}_{kU^-}(x)$ are the output of indicator functions that were 1 when the image $x$ was annotated, positive for class $k$ (unannotated) and negative for class $k$ (unannotated), respectively.

\subsubsection{Avoiding numerical underflow and balanced range normalization}
\label{sec:balancing}
To avoid numerical underflow and have a more uniform range for the output of models, Li et al.~\citep{cvpr2018} suggested to normalize the factors of the products in Equations~\ref{eq:output} to~\ref{eq:lossmil}, i.e., $\sigma(\gamma_{jk})$ and $1-\sigma(\gamma_{jk})$, to the range $\left[0.98,1\right]$. After running tests, we achieved better results by balancing this normalization, changing the range of each product factor to $[0.0056738^{(1/n_t)},1]$, where $n_t$ is the number of factors being multiplied. This range allows all products to have a similar expected range and keeps the same [0.98,1] range when $n_t=256$.

\subsection{Multi-task learning}
\label{sec:multitask}
Inspired by a work by Karargyris et al.~\cite{ibm}, we added another loss term to our method. With the intuition of giving more supervision to the representations calculated by encoder $E$ and, consequently, improving its representations, we added the task of predicting a high-resolution \gls*{ET} map for each label, performed with the help of decoder $D$, as shown in Figure~\ref{fig:method}c. From testing the network's performance, we modified the method in that class outputs were calculated according to Equation~\ref{eq:output} instead of adding fully connected layers as suggested by Karargyris et al.~\cite{ibm}.
Another difference in our method was that our decoder output had one channel for each of the ten labels in our classification task. The output of decoder $D$ could then provide estimations for the localization of abnormalities. In other words, the output of $D$ is an interpretability output: an alternative to the spatial activations or other interpretability methods, such as GradCAM~\cite{gradcam}. Decoder $D$ had an architecture with three blocks, each composed of a sequence of a bilinear upsampling layer, a convolution layer, and a batch normalization layer. The loss $L_T(x)$ added for this task is the pixel-level cross-entropy between the output of decoder $D$ and the label-specific \gls*{ET} map in the same resolution, $256\times256$, formulated as
\begin{equation}
\label{eq:pixelbce}
L_T(x) = -\mathbb{E}_{x \in X,\,k \in K}\left[\mathds{1}_{kA^+}(x)\mathbb{E}_{j\in \Gamma_x}\left[{\mathds{1}_{j \in B_{kx}}\text{log}\left(\sigma(\gamma_{jkx})\right)}+\mathds{1}_{j \notin B_{kx}}\text{log}\left(1-\sigma(\gamma_{jkx})\right)\right]\right],
\end{equation}
where $\mathds{1}_{kA^+}(x)$ is the output of an indicator function that was 1 when the image $x$ was annotated and positive for class $k$.
This loss was used to train both decoder $D$ and encoder $E$ and, as shown in Equation~\ref{eq:pixelbce}, was applied only for channels corresponding to positive ground-truth labels.

\subsection{Multi-resolution Architecture}
\label{sec:multires}
As shown in Figure~\ref{fig:multires}, for our encoder $E$ we adapted the Resnet-50~\citep{resnet} architecture by replacing its average pooling and last linear layer with two convolutional layers separated by batch normalization~\citep{batchnorm} and ReLU activation (CNN Block 5 from Figure~\ref{fig:multires}). To improve the results for labels with small findings in the original image, we modified the network such that spatial maps with $32\times 32$ resolution were used as inputs to CNN Block 5.

\subsection{Loss function}

The final loss function $L(\cdot)$, to be minimized while training, is given by 
\begin{equation}
\label{eq:lossfinal}
L(x) = L_I(x) + \lambda_{T} L_{T}(x),
\end{equation}
where $\lambda_{T}$ is a hyperparameter controlling the relative importance of $L_T(x)$. The classification output of our model only influences the $L_I(x)$ term.

\subsection{Datasets}

We used two datasets in our study. The REFLACX dataset~\citep{reflacxphysionet,reflacx1,physionet} provides \gls*{ET} data and reports from five radiologists for \glspl*{CXR} from the MIMIC-CXR-JPG dataset~\citep{physionet,mimicxrjpgphysionet,mimiccxrjpg}. Additionally, the REFLACX dataset contains image-level labels and radiologist-drawn abnormality ellipses, which can be used to validate the locations highlighted by our tested models. Except for the experiments described in Sections \ref{sec:labelerexperiments} and \ref{sec:extractionexperiments}, we used examples from Phase 3 from the REFLACX dataset. The MIMIC-CXR-JPG dataset, which contains patients who visited the emergency department of the Beth Israel Deaconess Medical Center between 2011 and 2016, was also utilized for its unannotated \glspl*{CXR} and image-level labels. Images from the MIMIC-CXR-JPG dataset were filtered using the same criteria as the REFLACX dataset: only labeled frontal images from studies with a single frontal image were considered. The test sets for both datasets were kept the same. A few subjects from the training set of the REFLACX dataset were assigned to its validation set so that around 10\% of the REFLACX dataset was part of the validation set. The same subjects were also assigned to the validation set of the MIMIC-CXR-JPG dataset. The train, validation, and test splits had, respectively, 1724, 277, and 506 images 
 for the annotated set and 187519, 4275, and 2701 images 
 for the unannotated set. The use of both datasets did not require ethics approval because they are publicly available de-identified datasets.

The sets of labels from the annotated and unannotated datasets were different. We decided to use the ten labels listed in Table~\ref{tab:labelresultsauc}. We provide, in Table \ref{tab:labelcorrespond}, a list of the labels from each dataset that were considered equivalent to each of the ten labels used in this study and, in Table \ref{tab:labelnumber}, the number of examples of each label present in the datasets.

\subsection{Labeler}
\label{sec:labelerexperiments}
The set of labels from the REFLACX dataset is slightly different from the ones provided by the CheXpert labeler. With the help of a cardiothoracic subspe\-cialty-trained radiologist, we modified the labeler to output a new set of labels. Modifications were also made to improve the identification of the already present labels after observing common mistakes on a separate validation set composed of 20\% of Phase 1 and Phase 2 from the REFLACX dataset. We adjusted rules for negation finding and added/adapted expressions to match and unmatch labels\footnote{The final set of rules can be found in our code repository at \url{https://github.com/ricbl/eye-tracking-localization}}.

\subsection{Location extraction}
\label{sec:extractionexperiments}
We tested several methods for extracting label-specific localization of abnormalities from the eye-tracking data. All methods involved the accumulation of fixations into heatmaps, with different starting and ending accumulation times, after extracting the label's mention time from the dictation. For a first stage of validation, the starting times we considered were:
\begin{itemize}
\item MAX(Start of mention sentence - TIME, Start of the previous sentence),
\item MAX(First mention in the sentence - TIME, Start of the previous sentence),
\item MAX(End of mention sentence - TIME, Start of the previous sentence),
\item start of first report sentence,
\item start of the previous sentence,
\item end of the previous sentence,
\item start of mention sentence,
\item start of the recording of data for that \gls*{CXR},
\end{itemize}
where TIME is a time delay assuming the values of 2.5s, 5.0s, and 7.5s. The end times we considered were:
\begin{itemize}
\item start of mention sentence,
\item end of mention sentence,
\item end of the first mention,
\item end of the last mention.
\end{itemize}
We tested all combinations between starting times and end times with a duration of 0s or more. We compared the extracted heatmaps with the validation hand-anno\-tated ellipses using the IoU metric with a validated threshold. After this validation, we finetuned, as a second stage of validation, the time delay by testing more times (0.5s, 0.75s, 1s, 1.25s, 1.5s, 1.75s, 2s, 2.5s, 3s, 3.5s, 4s, 4.5s, 5s).

\subsection{Validation and evaluation}

For our experiments\footnote{The code for our experiments can be found at https://github.com/ricbl/eye-tracking-localization
}, we used PyTorch 1.10.2~\citep{pytorch}. Hyperparameters commonly used for \gls*{CXR} classifiers were employed during training and were not tuned for any tested method. Models were trained for 60 epochs with the AMSGrad Adam optimizer~\citep{amsgrad} using a learning rate equal to 0.001 and weight decay of 0.00001. A batch size of 20 images was chosen for the use of GPUs with 16GB of memory or more. Images were resized such that their longest dimension had 512 pixels, whereas the other dimension was padded with black pixels to reach a length of 512 pixels. For training, images were augmented with rotation up to 45 degrees, translation up to 15\%, and scaling up to 15\%. The grid supervised by loss $L_I(x)$ had 1024 cells ($32\times32$). We used the max-pooling operation to convert the \gls*{ET} heatmap annotations to the same dimension. We thresholded the \gls*{ET} heatmaps at 0.15. This number was chosen after visual analysis of their histograms of intensities. We used $\lambda_A=3$ and $\lambda_T=300$ after validation of AUC and IoU values for our proposed method considering the following values: 0.3, 1, 3, 10, 30, and 100 for $\lambda_A$ and 0.1, 0.3, 1, 3, 10, 30, 100, 300, 1000, 3000 for $\lambda_T$. We trained models with five different seeds and report their average results and 95\% confidence intervals. Experiments were run in internal servers containing Nvidia GPUs (TITAN RTX 24GB, RTX A6000 48GB, Tesla V100-SXM2 16 GB). Each training run took approximately two to three days in one GPU.

As baselines, we evaluated a model trained without the annotated data (\textit{Unannotated}) and a model trained with data annotated by the drawn \gls*{GT} truth ellipses (\textit{Ellipse}). The ellipses were represented by binary heatmaps and were processed in the same way as the \gls*{ET} heatmaps. The loss function, \gls*{CNN} architecture, and training hyperparameters were the same for all methods. We did not include a cross-entropy classification loss baseline because it achieved lower scores than the presented methods. The best epoch for each method was chosen using the average \gls*{AUC} on the validation set. The best model heatmap threshold for each method and label was calculated using the average validation \gls*{IoU} over the five seeds, considering the full range of thresholds.

We evaluated our model (\textit{\gls*{ET} model}) and the two baselines by calculating the test \gls*{AUC} for image-level labels of the MIMIC-CXR-JPG dataset and the test \gls*{IoU} for localization of abnormalities, compared against the drawn \gls*{GT} ellipses. \Gls*{IoU} was calculated individually per positive label for all images with a positive label. We calculated three heatmaps for each label and input image: the output of decoder $D$, the spatial activations, and the output of the GradCAM method~\citep{gradcam}. We tested which heatmap had the best IoU validation results for each of the three reported training methods. We report results for the \textit{Ellipse} and \textit{\gls*{ET} model} using the output of decoder $D$ and for the \textit{Unannotated} model using the spatial activations. The heatmaps for each method were upscaled to the resolution of the \gls*{GT} ellipses using nearest-neighbor interpolation.

\section{Results}

\subsection{Labeler}

Labeler quality estimations, after modifications, are shown in Table~\ref{tab:labeler} and were calculated with the rest of the data from Phase 2, representing 80\% of the cases. Results were variable depending on the label, and misdetections should be expected when using this labeler.

\subsection{Location extraction}

For the first validation, the highest-scoring accumulated heatmaps used a starting time of MAX(Start of mention sentence - 2.5s, Start of the previous sentence) and an ending at the end of the last mention present in the sentence. For the second stage of the location extraction validation, when we tested delay times in a higher resolution, the time with the best IoU was 1.5s with an IoU of 0.233, justifying our approach as presented in Section~\ref{sec:extraction}.

\subsection{Comparison with baselines}

Results, averaged over all labels, are presented in Table~\ref{tab:groupresults}. The average \gls*{AUC} for the \textit{\gls*{ET} model} was not significantly different from the baselines. Regarding localization, the \gls*{IoU} values showed that training with the \gls*{ET} data was significantly better than training without annotated data and worse than training with the hand-labeled localization ellipses. Results for \gls*{AUC} and \gls*{IoU} of individual labels are presented in Tables~\ref{tab:labelresultsauc} and~\ref{tab:labelresultsiou}. \Gls*{AUC} was stable among all methods for almost all labels. 
Successful and unsuccessful heatmaps generated by our trained models are shown in Figure~\ref{fig:examples}.

When training the \textit{Ellipse} model with only 15\% of the annotated dataset, we achieved an \gls*{IoU} of .257 [.248,.266]. Therefore, given the IoU provided for the \textit{\gls*{ET} model} in Table~\ref{tab:groupresults}, the best estimation for the value of \gls*{ET} data is 15\%, i.e., around one seventh, of the value of the hand-annotated data.

\subsection{Ablation study}

We present in Table~\ref{tab:ablation} the results of an ablation study with each modification added to the original method from Li et al.~\citep{cvpr2018}, which is presented in Section~\ref{sec:lietal}, including the label specific heatmaps from Section~\ref{sec:extraction}, the balanced range normalization from Sections~\ref{sec:balancing}, the multi-resolution architecture from Section~\ref{sec:multires}, and the multi-task learning from Section~\ref{sec:multitask}. Table~\ref{tab:ablation} presents the elements added to the model in chronological order of addition to our project in its first five rows. We show an advantage from each modification for all methods. We also added a row with the removal of only the label-specific heatmaps to show that they had a big impact on the final \gls*{IoU}. Its removal caused a decrease of around 0.062 (24.2\%) on the \gls*{IoU}, reaching an \gls*{IoU} similar to the \textit{Unannotated} model.

\section{Discussion}

Other studies have applied \gls*{ET} data for localizing abnormalities and improving the localization of models. Stember et al.~\citep{mridictation} showed that radiologists looked at the location of a label when indicating the presence of tumors in MRIs. However, we use a less restrictive and more challenging scenario, with freeform reports and multiple types of abnormalities reported. Saab et al.~\citep{miccai2021} showed that, when gaze data are aggregated in hand-crafted features and used in a multi-task setup, the GradCAM heatmap of a model overlaps more often with the location of pneumothoraces. Li et al.~\citep{glaucoma} developed an attention-guided network for glaucoma diagnosis split into three sequential stages: a prediction of an attention map supervised by an ET heatmap, an intermediary classification network trained to refine the attention map through guided backpropagation, and a final classification network. Wang et al.~\citep{knee} performed osteoarthritis grading by enforcing the \gls*{CAM} heatmap to be similar to the \gls*{ET} map, allowing for uncertainty in the \gls*{ET} map. We tested adding this method to our loss, but no improvement was seen. All of these methods used \gls*{ET} datasets where radiologists focused on a single task/abnormality, making their \gls*{ET} data intrinsically label-specific and their setup distant from clinical practice. The method we propose uses dictations to identify moments when the radiologist looks at evidence of multiple abnormalities, allowing for the use of \gls*{ET} data collected during clinical report dictation. We used a multi-task formulation as part of our loss following Karargyris et al.~\citep{ibm}. However, even though they used a dataset where radiologists looked for several types of abnormalities, they showed only the impact of a single \gls*{ET} heatmap for all labels. Our study focuses on more complex uses of the \gls*{ET} data, with the generation of label-specific heatmaps.

Contrary to other works, such as Karargyris et al.~\citep{ibm} and Li et al.~\citep{cvpr2018}, we achieved no improvements in classification performance in our setup when applying a variety of localization losses to our model. One of the reasons we might achieve different levels of improvement from Karargyris et al.~\citep{ibm} is that we use a much larger dataset for training the model, weakly including most of the MIMIC-CXR-JPG dataset. The use of abundant unannotated data might reduce the impact of the annotated data on the final model. However, improvements in the ability to localize the abnormalities were still shown in our experiments.

The method we proposed for producing label-specific localization annotations from \gls*{ET} data and for training models to produce heatmaps that match the annotation  improved the interpretability of deep learning models for \glspl*{CXR}, as measured by comparing produced heatmaps against hand-annotated localization of abnormalities. We also showed that, in our setup, around seven \glspl*{CXR} with \gls*{ET} data provide the same level of efficacy in localization supervision as one \gls*{CXR} with expert-annotated ellipses, showing the potential of using this type of data for scaling up annotations. As shown by our ablation study, the use of label-specific annotations was essential to the added value of using of the \gls*{ET} data.

The \gls*{ET} data were relatively noisy, and the achieved IoU for our label-specific training heatmaps was relatively low, limiting the achieved IoU for our proposed method of using \gls*{ET} data for training. In future work, we will investigate other methods of extracting the localization information to reduce the noise in the data, including methods of unsupervised alignment. 

\section*{Conflict of Interest Statement}

The authors declare that the research was conducted in the absence of any commercial or financial relationships that could be construed as a potential conflict of interest.

\section*{Author Contributions}

R.B.L. wrote the manuscript, coded and conducted the experiments, and ran the analyses. 
J.S. participated in the study design and provided feedback for the manuscript. 
T.T. is the project PI, coordinating the study design, leading discussions about the project, and editing the manuscript.
All authors reviewed the manuscript.

\section*{Funding}
This research was funded by the National Institute Of Biomedical Imaging And Bioengineering of the National Institutes of Health under Award Number R21EB028367.

\section*{Acknowledgments}
Yichu Zhou participated in the modification and evaluation of the CheXpert labeler. This article has appeared in a preprint~\cite{preprint}.

\section*{Data Availability Statement}
The datasets analyzed for this study can be found in the PhysioNet repository at \url{https://www.physionet.org/content/reflacx-xray-localization/1.0.0/} and \url{https://www.physionet.org/content/mimic-cxr-jpg/2.0.0/}.

\bibliographystyle{Frontiers-Vancouver}
\bibliography{test}

\section*{Figure captions}
\setcounter{figure}{0}

\begin{subfigure}[H]
\setcounter{subfigure}{0}
    \centering
    \begin{minipage}[b]{0.35\textwidth}
        \includegraphics[width=\textwidth]{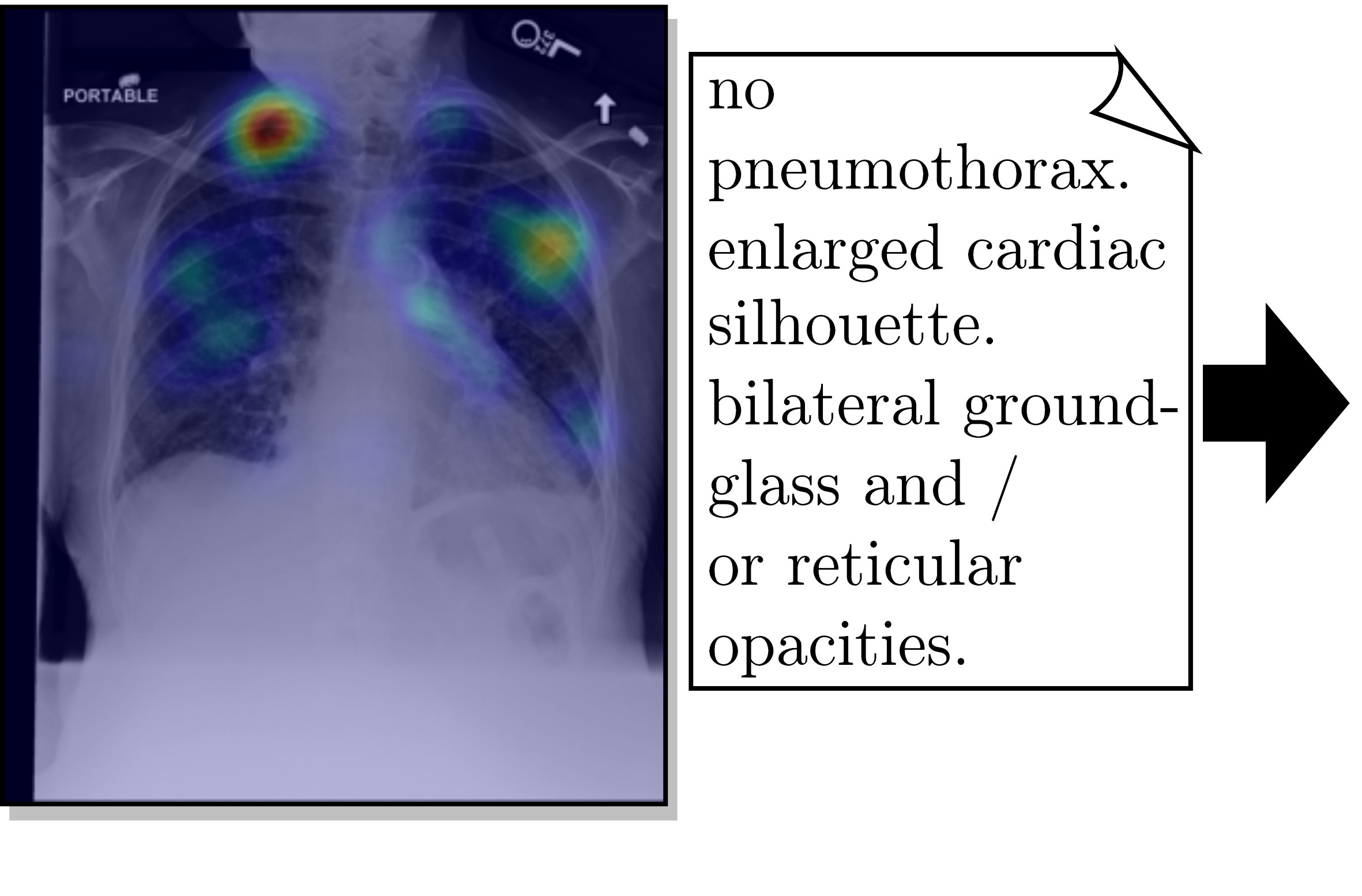}
        \caption{Original data.}
        \label{fig:Subfigure 1}
    \end{minipage}  
    \begin{minipage}[b]{0.35\textwidth}
        \includegraphics[width=\textwidth]{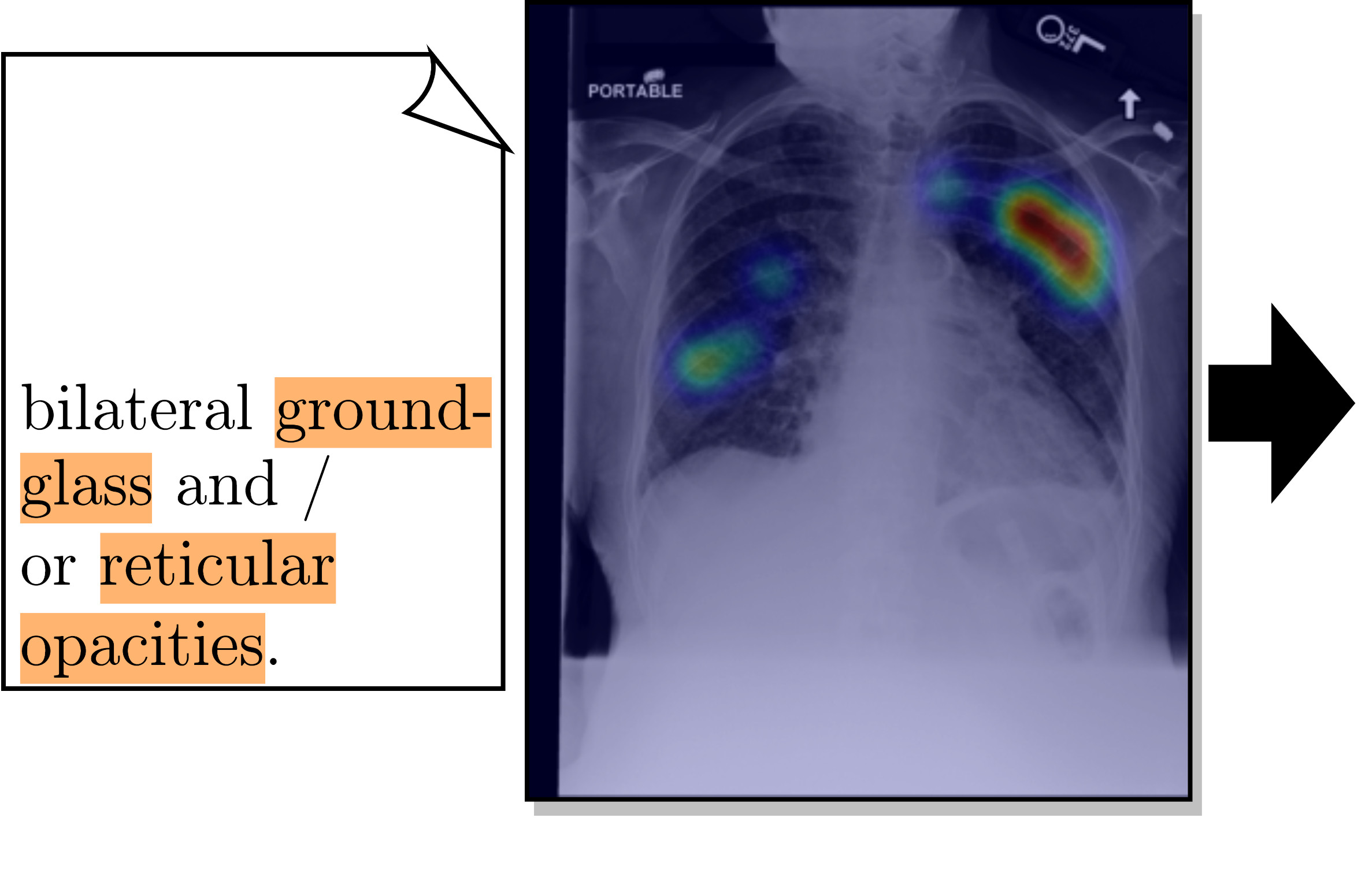}
        \caption{Label-specific heatmap.}
        \label{fig:Subfigure 2}
    \end{minipage}
    \begin{minipage}[b]{0.27\textwidth}
        \includegraphics[width=\textwidth]{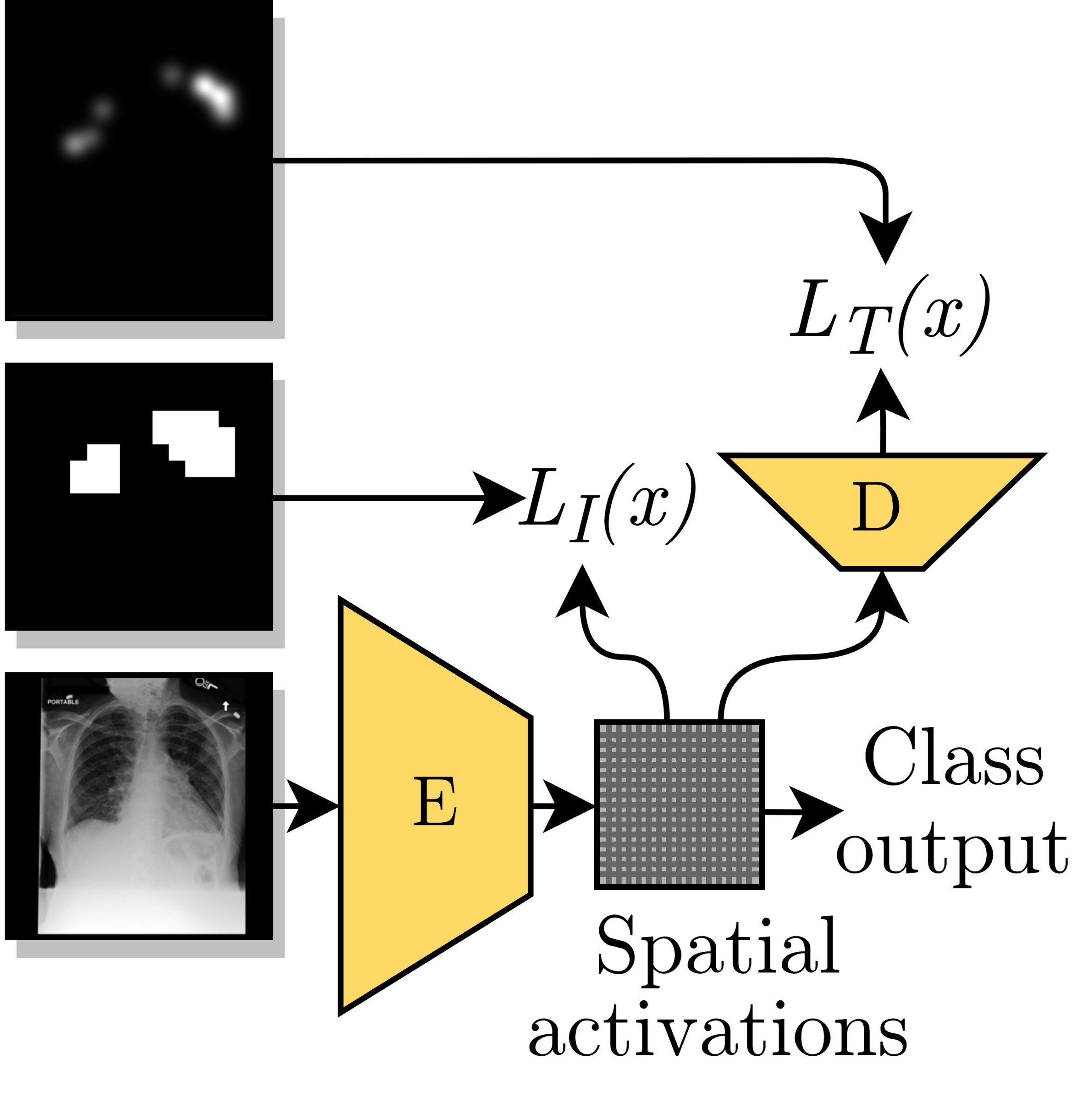}
        \caption{Training method.}
        \label{fig:Subfigure 3}
    \end{minipage}
    \setcounter{subfigure}{-1}
    \caption{Diagram of the use of ET data from a radiologist to train a CNN for improved localization. \textbf{(A)} The ET heatmap from the dictation of the full report over its corresponding CXR. \textbf{(B)} Label-specific ET heatmap for the label \textit{Opacity}. The keywords associated with this label, found by the adapted CheXpert labeler, are highlighted in orange. The listed sentence represents the timestamps from which fixations were extracted for generating the label-specific heatmap. \textbf{(C)} A representation of the employed loss function, which compares the extracted heatmap against an encoded spatial vector and a decoded version of it.
    }
    \label{fig:method}
\end{subfigure}



\begin{figure}[H]
\begin{center}
\includegraphics[width=\textwidth]{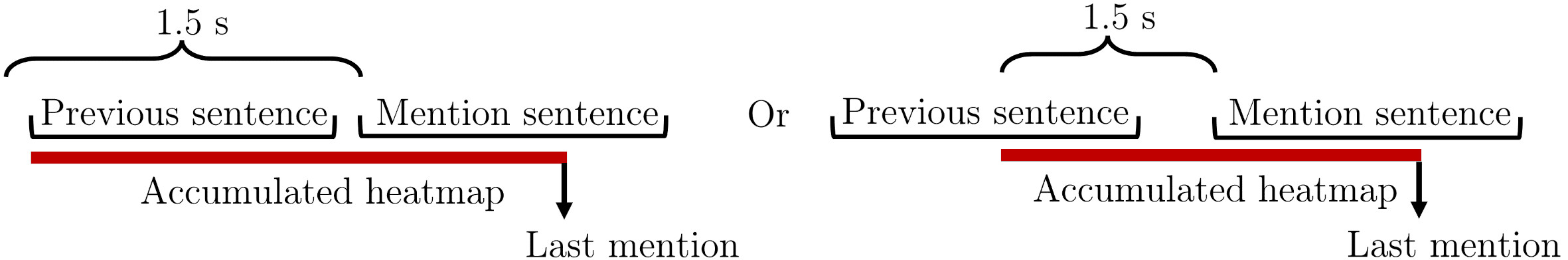}
\end{center}
\caption{Method of accumulation of fixations for generating heatmaps for each sentence that mentions at least one of the abnormality labels. The chosen fixations could be between the start of the previous sentence and the last mention of the current sentence or between 1.5 seconds before the start of the current sentence and the last mention of the current sentence, whichever has the shortest duration.}
\label{fig:15smethod}
\end{figure}

\begin{figure}[H]
\begin{center}
\adjustbox{minipage=0.90\linewidth,valign=t}{
\includegraphics[width=\textwidth]{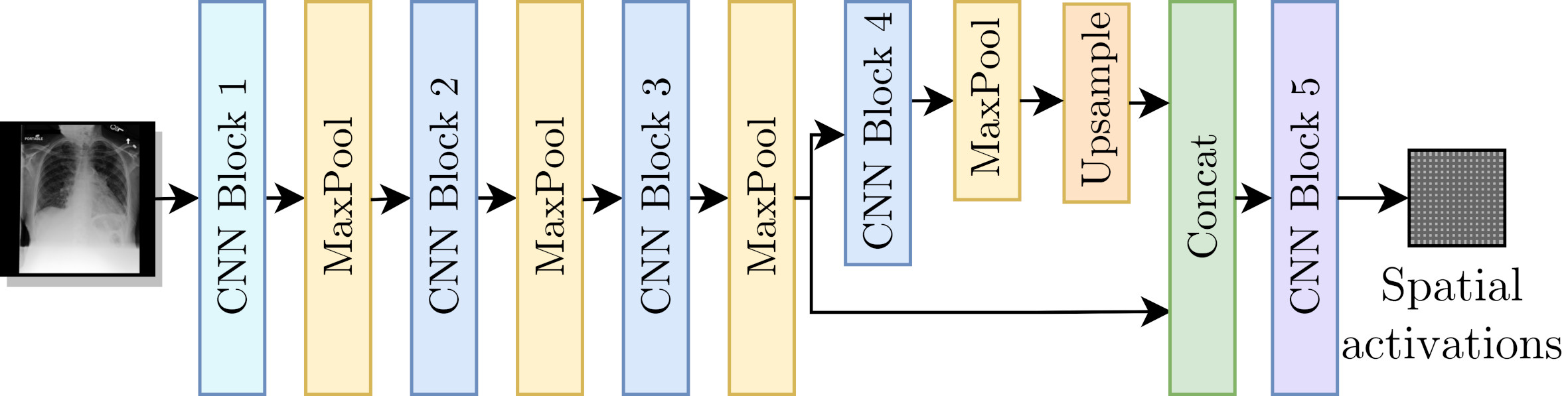}}
\centering  \adjustbox{minipage=0.05\linewidth,valign=t}{
}
\end{center}
\caption{Encoder $E$ as a modified Resnet-50 architecture to include the multi-resolution branches} \label{fig:multires}
\end{figure}

\begin{figure}[H]
\centering \adjustbox{minipage=0.90\linewidth,valign=t}{
\includegraphics[width=\textwidth]{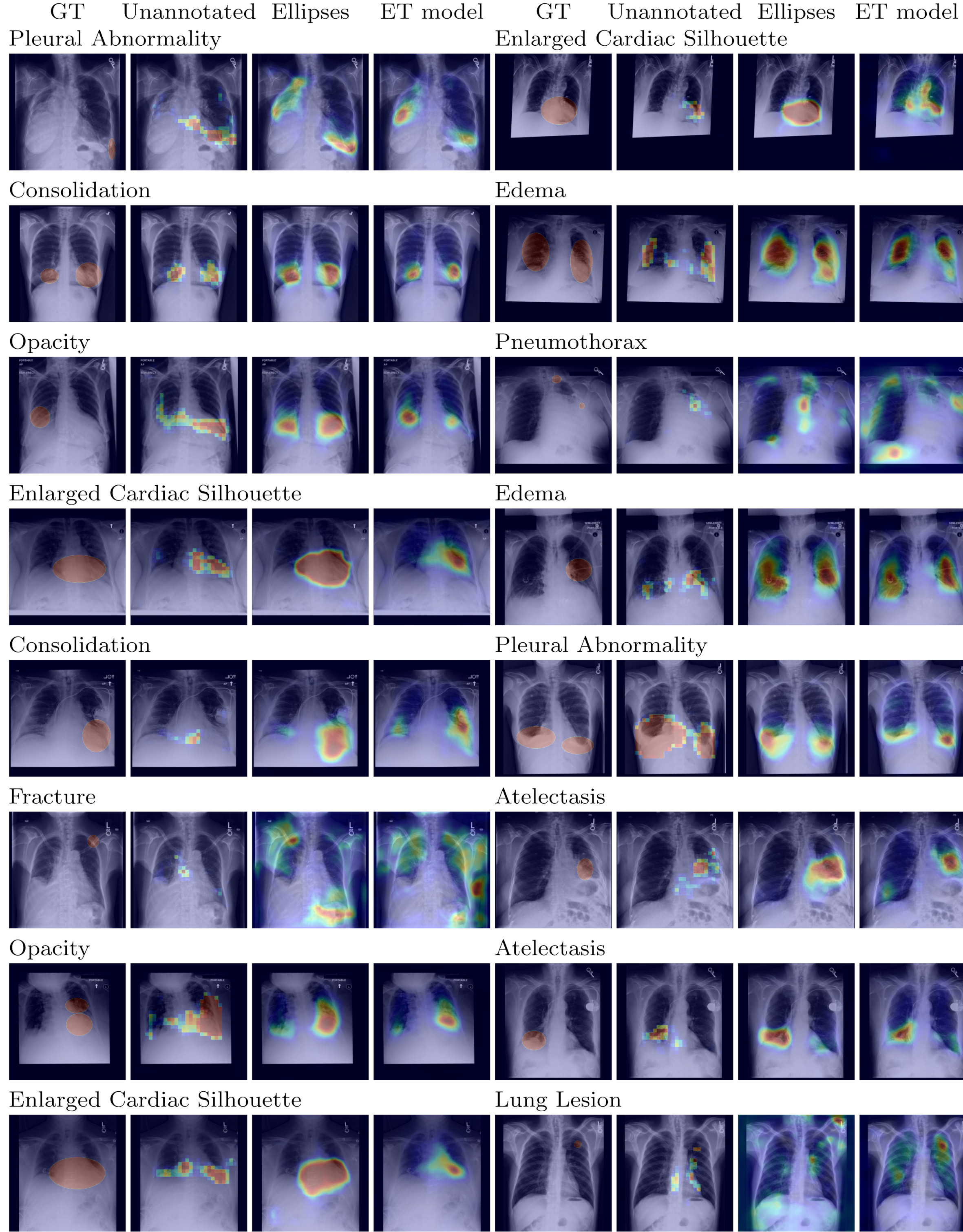}}
\caption{Localization output for the models for random test CXRs.
} \label{fig:examples}
\end{figure}


\section*{Tables}

\begin{table}[H]
\caption{Per-label AUC metric on the test set for the two baselines and our method.}\label{tab:labelresultsauc}
\centering
\begin{tabular}{|l|c|c|c|} 
								\hline
Label	&	\textit{Unannotated}	&	\textit{Ellipse}	&	\textit{ET model} (ours)	\\	\hline
AMC & 0.658 [0.654,0.662] & 0.665 [0.660,0.670] & 0.660 [0.655,0.665] \\ 
Atelectasis & 0.749 [0.744,0.754] & 0.751 [0.748,0.753] & 0.748 [0.746,0.751] \\ 
ECS & 0.768 [0.764,0.772] & 0.768 [0.765,0.771] & 0.765 [0.760,0.771] \\ 
Consolidation & 0.704 [0.697,0.711] & 0.709 [0.703,0.716] & 0.710 [0.706,0.713] \\ 
Edema & 0.839 [0.837,0.841] & 0.835 [0.833,0.837] & 0.838 [0.836,0.840] \\ 
Fracture & 0.714 [0.690,0.737] & 0.710 [0.696,0.724] & 0.714 [0.701,0.727] \\ 
Lung Lesion & 0.760 [0.747,0.774] & 0.751 [0.740,0.763] & 0.745 [0.738,0.752] \\ 
Opacity & 0.784 [0.782,0.786] & 0.785 [0.783,0.788] & 0.783 [0.779,0.787] \\ 
Pleural abnormality & 0.868 [0.865,0.870] & 0.869 [0.867,0.871] & 0.868 [0.865,0.871] \\ 
Pneumothorax & 0.825 [0.811,0.838] & 0.810 [0.799,0.820] & 0.819 [0.810,0.827] \\ 
 \hline

 \end{tabular}
\end{table}

\begin{table}[H]
\caption{List of labels that were grouped to form the labels from the analysis presented in this paper.}\label{tab:labelcorrespond}
\centering
\begin{tabular}{|p{4.5cm}|p{10cm}|}
\hline
Labels used in our models & Labels from public datasets (REFLACX and MIMIC-CXR) \\
\hline
Abnormal Mediastinal Contour (AMC) & abnormal mediastinal contour (AMC), enlarged cardiomediastinum \\
Atelectasis & atelectasis\\
Enlarged Cardiac Silhouette (ECS) & enlarged cardiac silhouette (ECS), cardiomegaly \\
Consolidation & consolidation\\
Edema & pulmonary edema, edema\\
Fracture & fracture, acute fracture \\
Lung Lesion & lung nodule or mass, lung lesion \\
Opacity & pulmonary edema, edema, lung nodule or mass, atelectasis, consolidation, groundglass opacity, interstitial lung disease, pneumonia, lung opacity \\
Pleural Abnormality & pleural abnormality, pleural other, pleural effusion \\
Pneumothorax & pneumothorax \\
\hline
\end{tabular}
\end{table}
\begin{table}[H]
\caption{Number of positive examples for each of the splits for both employed datasets: REFLACX (R) and MIMIC-CXR-JPG (M).}\label{tab:labelnumber}
\centering
\begin{tabular}{|l|r|r|r|r|r|r|}
\hline
Label	&	Train R	&	Val R	&	Test R	&	Train M	&	Val M	&	Test M	\\	\hline
Abnormal Mediastinal Contour	&	59	&	8	&	27	&	13551	&	351	&	289	\\	
Atelectasis	&	503	&	69	&	189	&	46981	&	1196	&	754	\\	
Enlarged Cardiac Silhouette	&	340	&	55	&	173	&	41681	&	1138	&	821	\\	
Consolidation	&	543	&	95	&	196	&	12311	&	352	&	253	\\	
Edema	&	222	&	39	&	124	&	33383	&	971	&	841	\\	
Fracture	&	34	&	5	&	27	&	4028	&	62	&	71	\\	
Lung Lesion	&	77	&	13	&	41	&	6015	&	114	&	106	\\	
Opacity	&	865	&	131	&	337	&	97640	&	2560	&	1861	\\	
Pleural Abnormality	&	486	&	69	&	205	&	50914	&	1426	&	1036	\\	
Pneumothorax	&	48	&	6	&	13	&	9653	&	258	&	96	\\

\hline
\end{tabular}
\end{table}

\begin{table}[H]
\caption{Results of the label detection with a modified version of the CheXpert labeler~\citep{chexpert}. AMC stands for \textit{Abnormal Mediastinal Contour} and ECS for \textit{Enlarged Cardiac Silhouette}.}\label{tab:labeler}
\centering
\begin{tabular}{|l|c|c|l|c|c|}
\hline
Label	&	Recall	&	Precision	&	Label	&	Recall	&	Precision	\\	\hline
AMC	&	0.67	&	0.73	&	Interstitial Lung Disease	&	0.75	&	0.27	\\	
Acute Fracture	&	1.00	&	1.00	&	Lung Nodule or Mass	&	0.50	&	0.50	\\	
Atelectasis	&	0.87	&	0.64	&	Pleural Abnormality	&	0.97	&	0.98	\\	
Consolidation	&	0.96	&	0.77	&	Pneumothorax	&	0.89	&	1.00	\\	
ECS	&	0.91	&	0.92	&	Pulmonary Edema	&	0.89	&	0.86	\\	
Groundglass Opacity	&	0.79	&	0.75	&		&		&		\\	

\hline
\end{tabular}
\end{table}

\begin{table}[H]
\caption{Results on the test set comparing the \textit{\gls*{ET} model} with the two baselines. AUC and IoU were averaged over the scores of all labels.}\label{tab:groupresults}
\centering
\begin{tabular}{|l|c|c|c|} 
								\hline
Method	&	\textit{Unannotated}	&	\textit{Ellipse}	&	\textit{ET model} (ours)	\\	\hline
AUC & 0.767 [0.763,0.771] & 0.765 [0.763,0.768] & 0.765 [0.763,0.767] \\ 
IoU & 0.201 [0.198,0.204] & 0.335 [0.330,0.339] & 0.256 [0.253,0.260]	\\	\hline

 \end{tabular}
\end{table}

\begin{table}[H]
\caption{Per-label IoU metric on the test set for the two baselines and our method.}\label{tab:labelresultsiou}
\centering
\begin{tabular}{|l|c|c|c|} 
								\hline
Label	&	\textit{Unannotated}	&	\textit{Ellipse}	&	\textit{ET model} (ours)	\\	\hline
AMC & 0.111 [0.100,0.122] & 0.297 [0.264,0.331] & 0.214 [0.187,0.242] \\ 
Atelectasis & 0.245 [0.241,0.250] & 0.385 [0.378,0.392] & 0.335 [0.320,0.350] \\ 
ECS & 0.386 [0.346,0.427] & 0.747 [0.743,0.751] & 0.379 [0.357,0.401] \\ 
Consolidation & 0.242 [0.235,0.249] & 0.380 [0.374,0.385] & 0.324 [0.314,0.334] \\ 
Edema & 0.314 [0.299,0.330] & 0.466 [0.460,0.472] & 0.401 [0.396,0.406] \\ 
Fracture & 0.012 [0.006,0.018] & 0.004 [0.004,0.004] & 0.007 [0.006,0.008] \\ 
Lung Lesion & 0.113 [0.104,0.123] & 0.203 [0.193,0.214] & 0.213 [0.194,0.231] \\ 
Opacity & 0.260 [0.255,0.264] & 0.387 [0.382,0.391] & 0.341 [0.336,0.347] \\ 
Pleural abnormality & 0.210 [0.202,0.218] & 0.297 [0.283,0.311] & 0.246 [0.240,0.251] \\ 
Pneumothorax & 0.114 [0.103,0.125] & 0.180 [0.173,0.187] & 0.103 [0.089,0.118] \\ 
 \hline

 \end{tabular}
\end{table}

\begin{table}[H]
\caption{IoU metric on the test set indicating the advantage of using each of the modifications to the multiple instance learning (MIL) method proposed by Li et al.~\citep{cvpr2018}. We tested different combinations of the following methods: label-specific heatmaps (LSH) (our contribution), balanced range normalization (BRN) (our contribution), multi-resolution architecture (MRA) (our contribution), and multi-task learning (MTL)~\citep{ibm}. }\label{tab:ablation}
\centering
\begin{tabular}{|c|c|c|c|c|c|} 
\hline
MIL & LSH & BRN & MRA & MTL	& IoU for \textit{ET model}	\\ \hline
$\checkmark$ & $\times$ & $\times$ & $\times$ & $\times$ & .165 [.162,.168]	\\
$\checkmark$ & $\checkmark$ & $\times$ & $\times$ & $\times$	& .180 [.176,.185]	 \\
$\checkmark$ & $\checkmark$ & $\checkmark$ & $\times$ & $\times$	&	.200 [.197,.203]	\\
$\checkmark$ & $\checkmark$ & $\checkmark$ & $\checkmark$ & $\times$	&	.218 [.216,.221]	\\
$\checkmark$ & $\times$ & $\checkmark$ & $\checkmark$ & $\checkmark$ & .194 [.192,.197] \\
$\checkmark$ & $\checkmark$ & $\checkmark$ & $\checkmark$ & $\checkmark$ & .256 [.252,.260] \\
 \hline 
 \end{tabular}
\end{table}

\end{document}